%% file: root.tex
\documentclass[letterpaper, 10 pt, conference]{ieeeconf}  

\IEEEoverridecommandlockouts                              

\usepackage{graphicx,float}
\usepackage{amsmath}
\usepackage{hyperref}
\usepackage{array}
\usepackage{flushend}
\usepackage{booktabs}
\usepackage{paralist}
\usepackage{multirow}
\usepackage{tabularx}
\usepackage{adjustbox}
\newcolumntype{Y}{>{\centering\arraybackslash}X}
\newcommand{\tabincell}[2]{\begin{tabular}{@{}#1@{}}#2\end{tabular}}

\DeclareMathOperator*{\argmin}{arg\,min}
\usepackage{xcolor}


\title{\LARGE \bf
Zero-shot Imitation Learning from Demonstrations for Legged Robot Visual Navigation
}

\author{Xinlei Pan$^{1,2}$, Tingnan Zhang$^{2}$, Brian Ichter$^{2}$, Aleksandra Faust$^{2}$, Jie Tan$^{2}$ and Sehoon Ha$^{2}$
\thanks{$^{1}$ University of California, Berkeley, CA, 94720, USA}
\thanks{$^{2}$ Robotics at Google, Mountain View, CA, 94043, USA}
\thanks{The research was conducted during Xinlei's internship at Google Brain. {\tt\small xinleipan@berkeley.edu, \{tingnan, ichter, faust, jietan, sehoonha\}@google.com}}
}

\begin{document}

\maketitle
\thispagestyle{empty}
\pagestyle{empty}

\begin{abstract}
Imitation learning is a popular approach for training effective visual
navigation policies. However, collecting expert demonstrations for legged robots is
challenging as these robots can be hard to control, 
move slowly, and cannot operate continuously for long periods of time.
In this work, we propose a zero-shot imitation learning framework 
for training a goal-driven visual navigation policy on a legged robot from
human demonstrations (third-person perspective), allowing for high-quality 
navigation and cost-effective data collection. 
However, imitation learning from third-person demonstrations 
raises unique challenges.
First, these demonstrations are captured from different camera perspectives, which we address via a feature disentanglement network~(FDN) that extracts perspective-invariant state features. 
Second, as transition dynamics vary between systems, we reconstruct missing action labels by either building an 
inverse model of the robot's dynamics in the feature space 
and applying it to the human demonstrations or developing a Graphic User Interface (GUI) to label human demonstrations.
To train a navigation policy we use a model-based imitation learning approach with FDN and action-labeled human demonstrations.
We show that our framework can learn an effective policy for a legged robot, Laikago, from human demonstrations in both simulated and real-world environments. 
Our approach is zero-shot as the robot 
never navigates the same paths during training as 
those at testing time.
We justify our framework by performing a comparative study.
\end{abstract}

\section{INTRODUCTION}


Legged robots have a great potential as universal mobility platforms for many real-world applications, such as last-mile delivery or search-and-rescue. However, visual navigation of legged robots can be considerably more challenging than wheeled robots due to the limited availability of legged robot navigation data: compared to over 1,000 hours for self-driving cars~\cite{yu2018bdd100k}, several kilometers~\cite{francis2019prmrl} and 32 years in simulation~\cite{chiang2019learning} for indoor mobile robot navigation. The reason of the difficulty for legged robot large scale data collection is that they are hard to control and operate continuously for a long time due to the hardware limitations.
This lack of data makes it difficult to deploy deep learning methods, such as reinforcement learning or imitation learning to the real world, especially when the robot are required to navigate towards a new goal unseen during training in the setting of zero-shot visual navigation.

We choose an imitation learning approach that obtains a navigation policy by mimicking human demonstrations, due to its data efficiency and data collection safety. However, for problems as complex as visual navigation, the amount of data required is outside of the scope of what is available for legged robots. Especially that the possible compositions of initial and goal states can be infinite while we can only collect limited data. 
In this work, our key insight is to mitigate the data collection issue by building a learning system that can learn to navigate from \emph{heterogeneous} experts--i.e., expert demonstrators that have different perspectives and potentially different dynamics. 
Our assumption is that these agents have better navigation capabilities than legged robots and are more readily available, thus alleviating the data collection bottleneck. 
Specifically, we focus on learning visual navigation from human agents. 


\begin{figure}
    \centering
    \includegraphics[width=0.48\linewidth,height=0.48\linewidth]{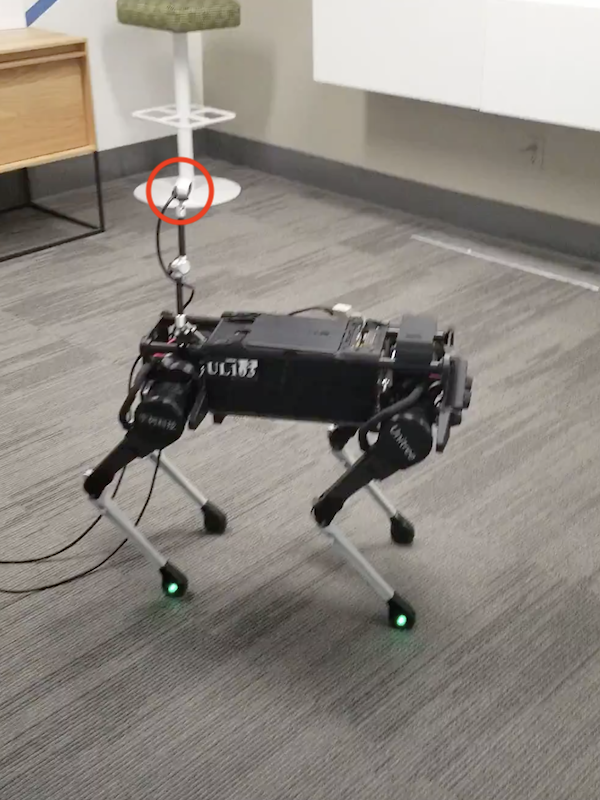}
    \includegraphics[width=0.48\linewidth,height=0.48\linewidth]{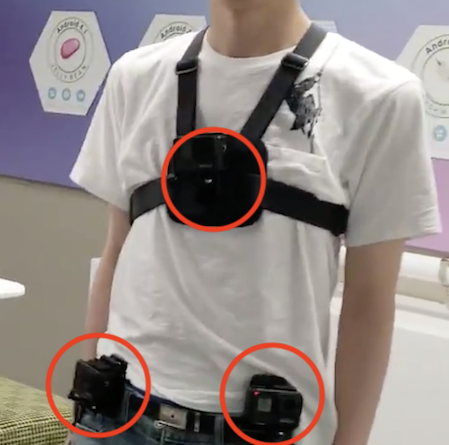}
    \caption{We develop a learning framework that trains a visual navigation policy for a legged robot (Left) from human demonstrations mounted with three cameras (Right). The red circles indicate cameras.}
    \label{fig:teaser}
\end{figure}

The idea of learning visual navigation from heterogeneous agents imposes new challenges. One of the main issues is the perspective shift between different agents' vision, because a robot may have different camera positions and viewing angles from other robots or humans. Directly transferring policies learned on human perspective demonstrations can result in domain shift problems. Additionally, in some cases, the demonstrations only contain raw state sequence, and do not contain action labels.
We thus need an effective planning module that finds the optimal actions solely based on raw images, without any additional information about the robot and surroundings.

In this work, we propose a novel imitation learning framework that trains a goal driven visual navigation policy for a legged robot from human demonstrations. A human provides navigation demonstrations as videos that are recorded by multiple body-mounted cameras. We extract relevant state features from the temporally-aligned multi-perspective videos by training a feature disentanglement network (FDN), which disentangles state related features from perspective related features. FDN achieves such disentanglement by training with our proposed cycle-loss, that composing disentangled features should be able to generate images with correspondence to the features. 
We consider two approaches for labeling demonstrations with robot-compatible actions, either via an efficient human labelling GUI or through a learned inverse dynamics model.
We then take a model-based imitation learning approach for training a visual navigation policy in the learned latent feature space. Our proposed approach is zero-shot in that the robot training data does not include the testing tasks and thus has to infer the policy from the human demonstrations.

We demonstrate our proposed framework in both simulated and real environments. Our framework can train effective navigation policies to guide a robot from the current position to the goal position described by a goal image. We also validate the feature disentanglement network by comparing the prediction of the perspective-changed images to the ground truth images. In addition, we analyze the performance of the proposed framework by conducting a comparative study and comparing with some baseline algorithms for feature learning and imitation learning. We observe that our approach achieves similar performance with perspective changes to that of the oracle imitation learning method without perspective change.
\vspace{-0.08in}

\section{Related Work}
\textbf{Robot Visual Navigation.}
Robot visual navigation is a fundamental task for mobile robots, such as legged robots. Traditional approaches such as simultaneous localization and mapping (SLAM) first constructs a map of the environment and then plan paths~\cite{bonin2008visual,konolige2010view,kerl2013dense}. 
However, these methods sometimes require a robot to navigate and gradually map the environment. Though these methods may work in normal navigation case, they may struggle in our case where the robot has to learn from human demonstrations of different perspectives and transition dynamics. 
Other approaches use learning to enable visual navigation through either imitation learning (next paragraph) or reinforcement learning (RL). RL based approaches learn to navigate given a reward function, either learned
from demonstration~\cite{ziebart2008maximum,fu2019language} or defined manually by human expert~\cite{wang2019reinforced}. Most existing work on visual navigation with reinforcement learning is done in simulation~\cite{fang2019scene, zhu2017target}; a few are done on real robots~\cite{pathak2018zero, gupta2017cognitive}. 
These approaches are limited in the legged robot case by requiring actual trial-and-error on real robots, which can be time intensive and dangerous as collisions on legged robots can easily damage themselves and the environment.

\textbf{Learning from demonstrations.}
Imitation learning~\cite{pan2018agile, liu2018imitation,ross2011reduction} learns a policy given labeled expert trajectories, such as imitating a goal driven navigation policy~\cite{zhu2017visual}, and conditional imitation learning~\cite{codevilla2018end}. As mentioned previously, imitation learning requires large quantities of labeled data that are not practical for legged robots.
The data could come from either on-robot demonstration such as imitating autonomous driving policy~\cite{rhinehart2018deep} or from human observation such as third-person imitation learning~\cite{stadie2017third}, learning by translating context~\cite{liu2018imitation} and using time-contrastive network (TCN) to learn a reward function~\cite{sermanet2018time}.
Though learning with on-robot data is effective, it is very labor intensive to collect large scale datasets for many robots, and some of them may require special training to use. Learning from human demonstrations of different perspectives is natural to mimic the way humans (e.g. children) learn to perform many control tasks by watching others (experts) performing the same tasks~\cite{meltzoff1999born}. However, the perspective shift between a human and robots is non-trivial. In this approach, we propose a novel feature extraction framework to solve this problem. In addition, our work is related with model-based reinforcement learning~\cite{pan2019semantic} and model-based imitation learning~\cite{srinivas2018universal}. Our imitation learning framework is similar to that of the universal planning  network~\cite{srinivas2018universal} (UPN), but differs in that we perform the model learning and planning in our learned feature space, rather than in the raw pixel space. Imitation learning on visual navigation from human video has been explored in~\cite{kumar2019learning}, where they propose to train an inverse dynamics model to learn an action mapping function from robot dynamics to human video. While their work focuses on learning subroutines for navigation, our work focuses on learning a perspective-invariant feature space that is suitable for path planning and model-based imitation learning. Our work could be combined with their contributions to improve the performance of visual navigation.

\textbf{Feature Disentanglement.} General feature disentanglement involves a broad spectrum of related works. Most works are done in image-to-image translation tasks such as~\cite{jha2018disentangling} and~\cite{lee2018diverse}, where they apply a similar cycle-consistency loss to achieve image translations across different domains. We propose a similar model for feature disentanglement that only uses the temporally aligned videos for feature learning, without additional supervision.
\vspace{-0.08in}

\section{Problem}
We consider learning goal-driven visual navigation policy on a legged robot from human demonstrations. In our work, a human expert mounts $N$ cameras on the body and walks in the training environment. Each demonstration yields a sequence of images $I^{1\cdots N}_{1\cdots T} \in \mathcal{I}$ with the perspective index (superscript) and time index (subscript). We assume that the images with the same time indices are captured at the same human state (their position in 2D and orientation). 

The robot’s observation space at time $t$ is defined by an image from the robot’s perspective $I^{robot}_t \in \mathcal{I}$. The action space consists of five discrete actions $a \in \mathcal{A}$: going forward, going backward, turning left, turning right and staying in place. Each action only provides high-level control over the robot while low-level motor torques on legs are computed by a traditional Raibert controller~\cite{raibert1986legged}. A goal image $I_{g}^{human}$ is specified from the human’s perspective. Therefore, the policy $\pi:(I^{robot}_t, I_g^{human})\rightarrow a$ maps the robot's observation $I^{robot}_t$ and the specified goal $I_g^{human}$ to the action $a$.
\vspace{-0.08in}

\section{Method}

\begin{figure}
    \centering
    \vspace*{0.1in}
    \includegraphics[width=0.9\linewidth]{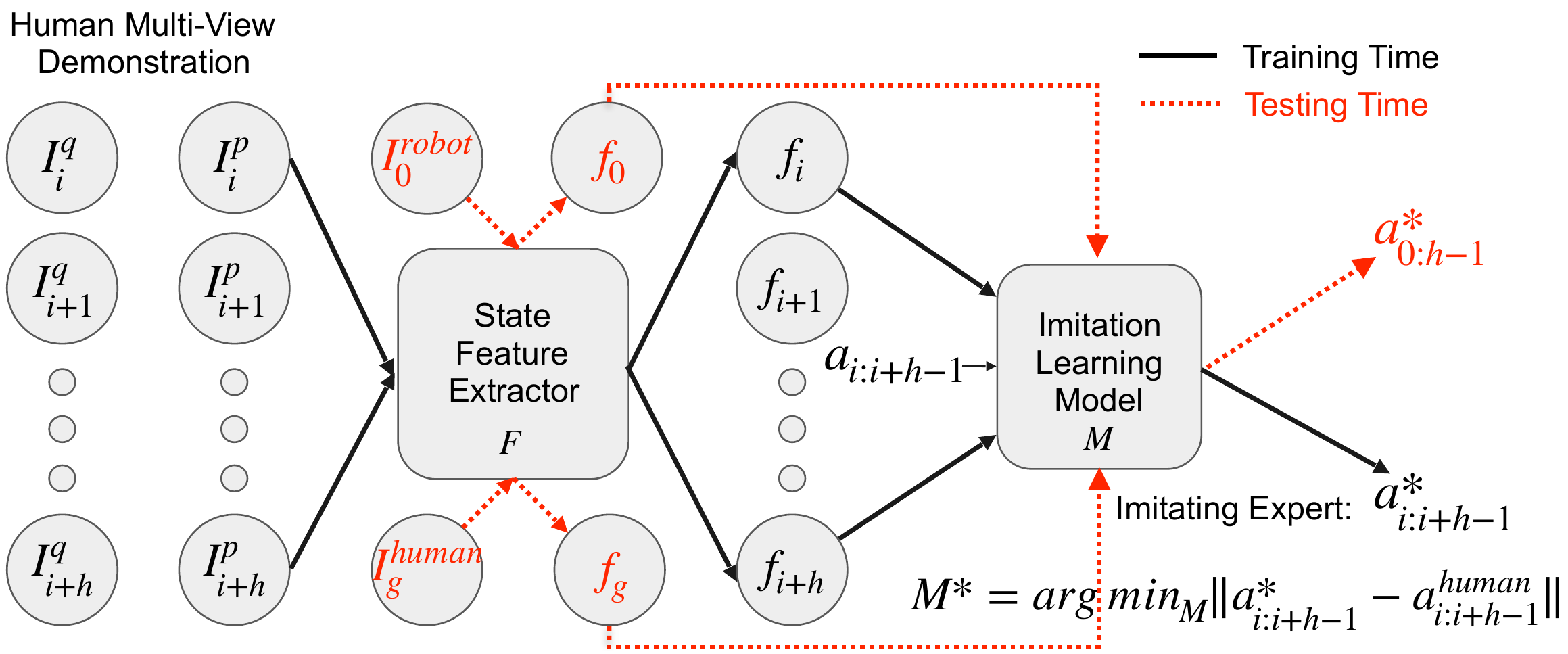}
    \caption{Overview of the proposed framework. 
    The feature extractor takes as inputs human demonstration trajectories from multiple perspectives (indicated by $p$ and $q$) and learns a perspective-invariant state feature extractor.
    The imitation learning model takes in the extracted feature of human demonstration data and learn to imitate the human action sequence $a^{human}_{i:i+h-1}$. 
    During testing (indicated by red dash line), start image $I_0^{robot}$ and goal image $I_g^{human}$ are fed into the state feature extractor and the imitation learning model takes in both the start feature $f_0$ and goal feature $f_g$ and optimizes an action sequence to minimize the difference between final state feature and goal state feature.  
    }
    \label{fig:systemdiagram}
\end{figure}

This section introduces a zero-shot imitation learning framework for visual navigation of a legged robot. We first introduce our feature disentanglement network (FDN) that extracts perspective-invariant features from a set of temporally-aligned video demonstrations. Then we present the imitation learning algorithm that trains a navigation policy in the learned feature space defined by FDN. Figure~\ref{fig:systemdiagram} gives an overview of the framework. 

\subsection{Feature Disentanglement Network}

\begin{figure}
    \centering
    \includegraphics[width=0.9\linewidth]{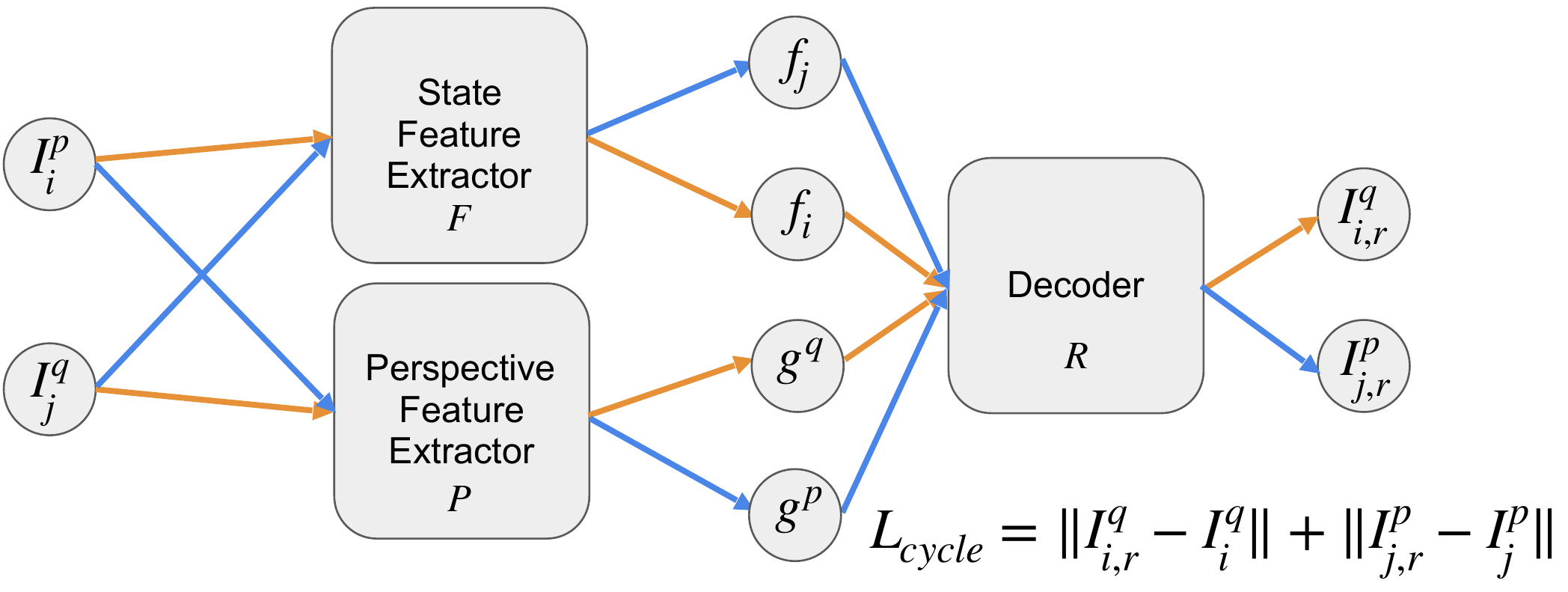}
    \caption{Diagram of the feature disentanglement network (FDN). FDN is composed of three sub-networks, the state feature extractor $F$, the perspective feature extractor $P$ and the decoder $R$. Given images of different states (indicated by $i,j$) and of different perspectives (indicated by $p,q$), the network first extracts and separates state/perspective information, then composes them together to generate another image that corresponds to the input state and perspective features. The blue lines indicate the feed-forwarding path to generate $I_{j,r}^p$ and the yellow lines for $I_{i,r}^q$.}
    \label{fig:fdn}
\end{figure}

We design a feature disentanglement network (FDN, Figure~\ref{fig:fdn}) to perform feature disentanglement from visual inputs. More specifically, the FDN tries to separate state feature from perspective feature, which is necessary for imitation learning between the heterogeneous agents. The network is composed of two parts: the state feature extractor $F_{\theta}$ with parameters $\theta$, which extracts state-only feature from the input; and the perspective feature extractor $P_{\phi}$ with parameters $\phi$, which extracts perspective-only feature from the input. For simplicity, we drop the parameters of the functions unless necessary. 

Denote the entire human demonstration dataset as $\mathcal{D}= \{I_i^p\}_{i=1:T}^{p=1:N}$ where $T$ is the total length and $N$ is the total number of perspectives. 
For a given image input $I_{i}^p$, both networks extract one part of information from the visual input: 
    $f_i = F(I_{i}^p), 
    g^p = P(I_{i}^p),$
where $f_i \in \mathcal{F}$ and $g^p \in \mathcal{G}$ are the corresponding state features and perspective features, respectively.
For training FDN, we learn an image reconstructor $R_{\psi}$ with parameters $\psi$ that takes in $f_i$ and $g^p$ and reconstructs an image corresponding to the same state specified by $f_i$ and the same perspective specified by $g^p$:
    $I^p_{i,r} = R_{\psi}(F(I^p_i)), P(I_{i}^p)),$
where the subscript $r$ denotes reconstructed image.
For any two images $I_i^p$, $I_{j}^q$ that correspond to different state features $f_i,f_j$ and different perspective features $g^p, g^q$, we define the cycle-loss function of training the feature extractor as:
   $ L_{cycle}(I^p_i, I_{j}^q, \theta, \phi, \psi) = \|I_{i}^q-R_{\psi}(F_{\theta}(I_{i}^p), P_{\phi}(I_{j}^q))\|+\|I_{j}^p-R_{\psi}(F_{\theta}(I_j^q), P_{\phi}(I_i^p))\|.$
Assuming access to temporally aligned images from multiple perspectives, the feature extractor will learn to extract state related information only in $F$ and learn to extract perspective information only in $P$. 
The total loss function for training FDN can be summarized by the following equation:
   $L_{total}(\theta, \phi, \psi) = \sum_{\forall i, j, p, q}L_{cycle}(I_i^p, I_j^q, \theta, \phi, \psi).$

We train FDN by randomly sampling two images from the multi-perspective data.
We use the CycleGAN~\cite{zhu2017unpaired} encoder as the backbone of the feature extractor and convert the last layer output as a flattened $d$ dimensional vector. The decoder or the image generator is inherited from CycleGAN decoder. 
The Swish activation function ~\cite{ramachandran2017swish} is used through the network when necessary.

\subsection{Imitation Learning from Human Demonstrations}

\begin{figure}
    \centering
    \vspace*{0.1in}
    \includegraphics[width=0.9\linewidth]{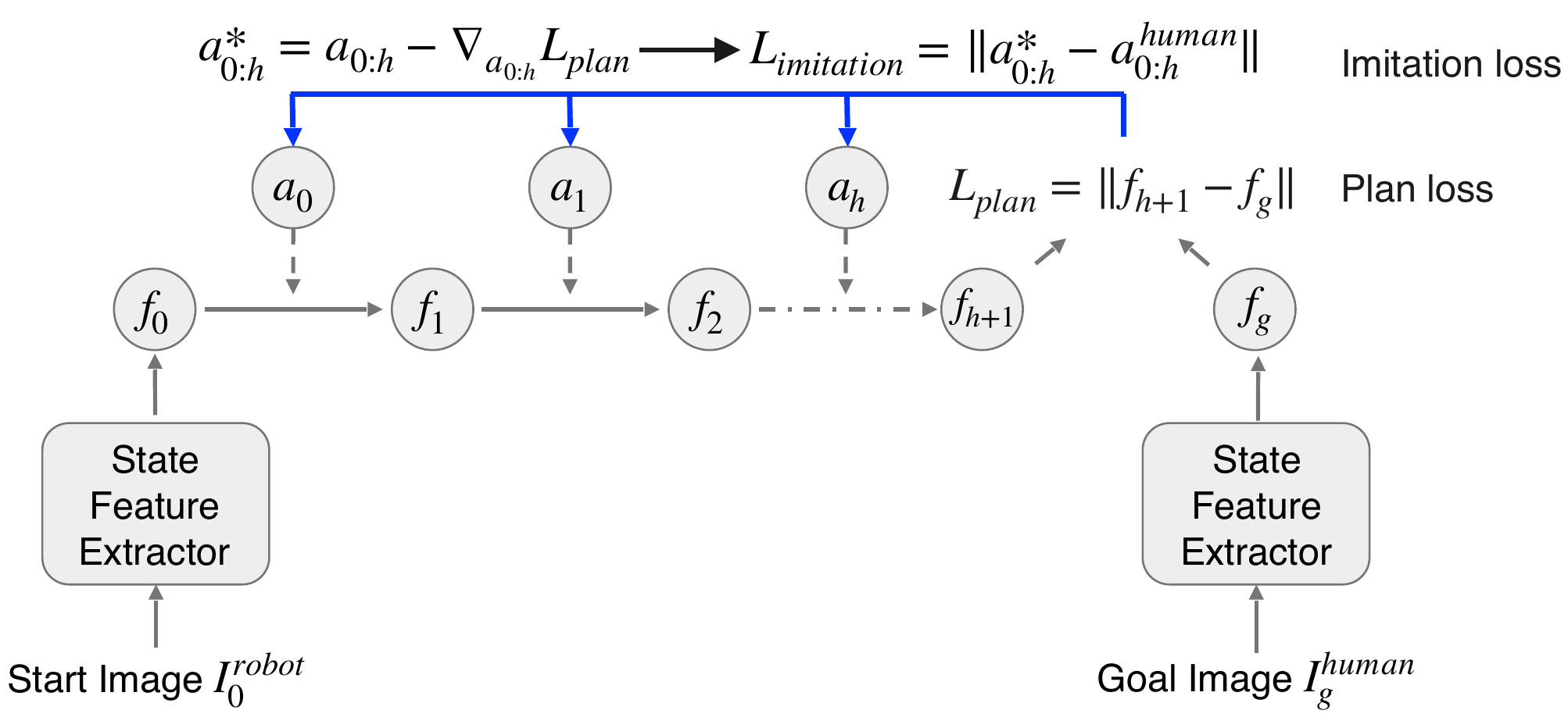}
    \caption{Diagram of the model-based imitation learning. The model takes in randomly initialized action sequence $a_{0:h}$ and predicts future state latent representations. It optimizes the action sequence to minimize the difference between the predicted final state $f_{h+1}$ and the goal state feature representation $f_g$ (gradient flow indicated by blue line). Model parameters are updated by minimizing the imitation loss.
    }
    \label{fig:imitation}
\end{figure}

Inspired by the Universal Planning Network [UPN], we train the model-based imitation learning network (Figure~\ref{fig:imitation}) in the latent feature space $\mathcal{F}$.
We process the given human demonstration data $\mathcal{D}$ into a sequence of the features $\{f_0, f_1, \cdots, f_{n}\}$ by applying the trained FDN to the data.
We label the robot-compatible actions  $\{a_0, a_1, \cdots, a_{n-1}\}$ by training an inverse dynamics model or using a developed GUI to manually label actions. The inverse dynamics model (IDM) takes in FDN state feature extractor processed images that is temporally consecutive and predicts the robot action that completes the transition. To get one episode of robot data for training IDM, we randomly start the robot and walk the robot in the environment until collision or the number of steps exceeds 30. We collect multiple episodes of such data. 


We define a model $M$ that takes in the current observation's feature encoding $f_0$, and a randomly initialized action sequence $a_{0:h}=\{a_0, a_1, \cdots, a_{h}\}$, where $h+1$ is the prediction horizon of the model, and predicts future states' feature representation $f_1,\cdots, f_{h+1}$.
Then we update the action sequence by performing gradient descent on the following plan loss:
\begin{equation}
\label{eq:huber}
   a^*_{0:h}
 =\argmin_{a_{0:h}}L_{H}( M(F(I_0^{robot}),a_{0:h}), F(I_g^{human})),
\end{equation}
which minimizes the difference between the predicted final future state feature $M(F(I_0^{robot}),a_{0:h})$ and the given goal state feature $F(I_g^{human})$. Here we use the superscript $robot$ to explicitly point out that $I_0^{robot}$ is from the robot's perspective while the superscript $human$ means $I_g^{human}$ is from the human's perspective.
We use the Huber loss~\cite{huber1992robust} ($L_H$) to measure the difference between the predicted feature and goal feature as it was used in~\cite{srinivas2018universal}.
Then given the human demonstrator's 
expert action sequence $a^{human}_{0:h}$, we optimize the model parameters so as to imitate the expert behavior:
\begin{equation}
    M^* = \argmin_{M}\|a^*_{0:h}-a^{human}_{0:h}\|,
    \label{eq:6}
\end{equation}
the loss function above could be a cross entropy loss when the action space is discrete. Once we train the model $M$, Eq. (\ref{eq:huber}) implicitly defines the policy $\pi$. At each time step, the policy replans the entire action sequence and only executes the first action, which is similar to the way model predictive control (MPC)~\cite{lenz2015deepmpc} does. When training the imitation learning model, the prediction horizon can change, and it depends on the number of expert steps between the start and goal state, a mask is applied on Equation~\ref{eq:6} to only imitate the corresponding action sequence. This is similar to the way UPN~\cite{srinivas2018universal} trains the policy. More details can be found in~\cite{srinivas2018universal}.
\vspace{-0.08in}

\section{Experiments and Analysis}
We design our experiments to investigate the following questions. 1) Is the proposed feature disentanglement network able to disentangle features? 2) Can the proposed model-based imitation learning find an effective action plan in the learned feature space? 3) How does our approach compare to other imitation learning methods? 

\subsection{Experiments and Results}
\begin{figure}
    \centering
    \vspace*{0.05in}
    \begin{tabularx}{\linewidth}{Y Y}
    \tabincell{c}{NavWorld} & \tabincell{c}{OfficeWorld} \\
    \includegraphics[width=0.5\linewidth, height=0.4\linewidth]{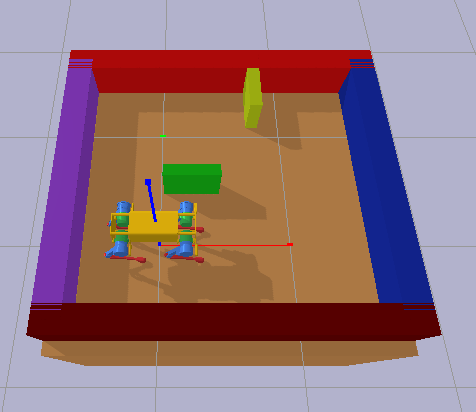} &
    \includegraphics[width=0.5\linewidth, height=0.4\linewidth]{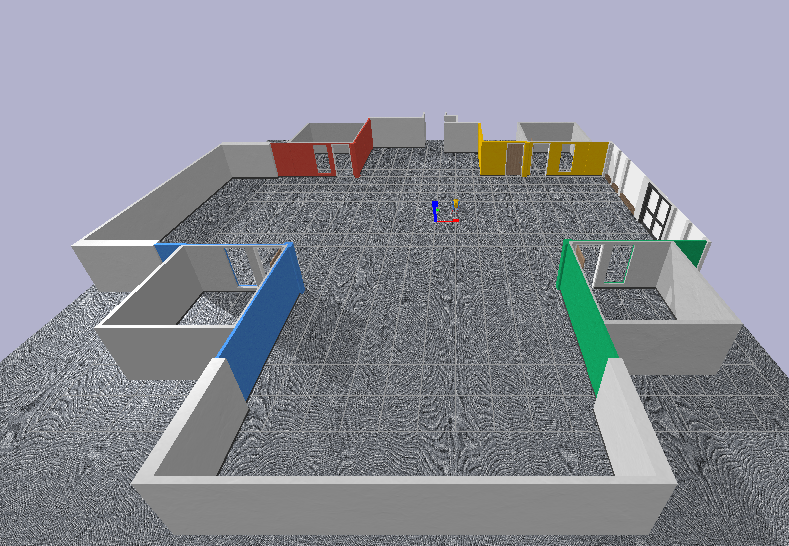}
    \end{tabularx}
    \caption{Simulation environments. Left: NavWorld ;
    Right: OfficeWorld.
    }
    \label{fig:env}
\end{figure}
\paragraph{Environment Setup and Data Collection}
We select Laikago from Unitree~\cite{laikago} as the real world robotic platform to evaluate the proposed framework. For simulation, we develop two simulation environments using PyBullet~\cite{coumans2016pybullet} (Figure~\ref{fig:env}), one is called Navworld, the other OfficeWorld. The latter one has more complex texture than the previous one, and the space is larger.
We put a simulated Laikago in both simulation environments.
The robot is $60$cm tall and we mount the camera $30$cm above the body (See Figure~\ref{fig:teaser} left). The frames are down-sampled to $128 \times 128$ pixels. The  discrete actions are defined as running a walking controller with a constant linear and angular velocity for $1$ second. As a result, a robot can move for about $0.2$m, which is near one-third of its body length, and can turn left or right for about $30$ degrees in the NavWorld environment and $90$ degrees in the OfficeWorld environment. We choose a small room of size $4\times 4 m^2$ for training and testing purposes for real robots. For simplicity, we assume kinematic movements. For all environments, we consider the task as given a goal image from another perspective, the robot needs to navigate towards the goal within limited number of steps (slightly longer than the optimal path). An episode terminates when the robot reaches the goal or when the maximum allowed steps are taken.

In our experiments, each human demonstration is a first-person view navigation video (human perspective). In the real world case, we collect temporally-aligned multi-perspective data by mounting three Go-Pro cameras on the person (Figure \ref{fig:teaser} right) and let the person navigate certain paths. 
The reason to mount the three cameras in these positions is to get perspectives of different heights and viewing angles that can interpolate the robot's perspective. The videos are downsampled to $128 \times 128$ pixels. We obtain multiple video clips with the same state sequence but different perspectives and extract perspective-invariant features. In total we collected 25 demonstration trajectories, each of length 20 steps in the real world environment for $10$ minutes. 
In simulation, we automatically generate demonstrations using a path-planning algorithm~\cite{dijkstra1959note} on randomly sampled start and goal locations. We collect 500 demonstration trajectories, each of length 20 steps. To improve the data efficiency, we perform data augmentation by replaying the video and reversing the time order both for the simulated and the real data. We add in augmented stay in place demonstration sequences by repeating randomly sampled observations for 20 steps.

In our framework, we need to obtain robot-compatible action labels of human demonstrations since they have different dynamics. In simulation, we trained an inverse dynamics model that takes in two consecutive images processed by FDN and predicts the robot action that completes the transition. Then we use the trained inverse dynamics model to label the expert demonstration. In the real world experiment, since the robot trajectory data especially the actions contain significant noise, we develop a GUI that allows us to label human actions manually within a short amount of time. Note that this work's focus is not on action labeling. In addition, the manually labeled action is only a rough estimation of where the robot is going, and it may still contain noise: for example, when the robot is staying in place, it may still move around a little bit due to drifting. 

\paragraph{Training}
We train the FDN and the imitation learning model both in the simulated environments and on the real robot. Additionally, we train the inverse dynamics model in the simulation for automatic human action labeling. For all experiments, we use the Adam optimizer~\cite{kingma2014adam} with a learning rate of 0.0035 and batch size of 32, and we set the feature dimension $d=256$. We evaluate the success (reaching the goal) rate of our experiments in simulation by comparing the robot's state (location and orientation) to the goal state, and the success rate on real robot by human visual evaluation.

\begin{figure}[ht]
    \centering
    \begin{tabularx}{\linewidth}{Y Y Y Y}
    \tabincell{c}{$I_i^p$} & \tabincell{c}{$I_j^q$} & \tabincell{c}{$I_i^q$ (target)} & \tabincell{c}{$I_{i,r}^q$ (pred.)} \\
    {\setlength{\fboxsep}{0pt}%
    \setlength{\fboxrule}{0.5pt}
    \fbox{\includegraphics[width=0.8\linewidth]{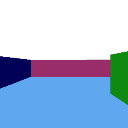}}} &
    {\setlength{\fboxsep}{0pt}%
    \setlength{\fboxrule}{0.5pt}
    \fbox{\includegraphics[width=0.8\linewidth]{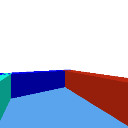}}}&
    {\setlength{\fboxsep}{0pt}%
    \setlength{\fboxrule}{0.5pt}
    \fbox{\includegraphics[width=0.8\linewidth]{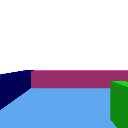}}}&
    {\setlength{\fboxsep}{0pt}%
    \setlength{\fboxrule}{0.5pt}
    \fbox{\includegraphics[width=0.8\linewidth]{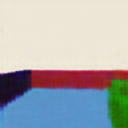}}} \\
    {\setlength{\fboxsep}{0pt}%
    \setlength{\fboxrule}{0.5pt}
    \fbox{\includegraphics[width=0.8\linewidth]{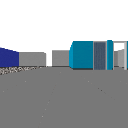}}}&
    {\setlength{\fboxsep}{0pt}%
    \setlength{\fboxrule}{0.5pt}
    \fbox{\includegraphics[width=0.8\linewidth]{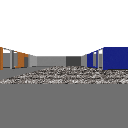}}}&
    {\setlength{\fboxsep}{0pt}%
    \setlength{\fboxrule}{0.5pt}
    \fbox{\includegraphics[width=0.8\linewidth]{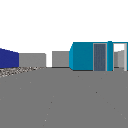}}}&
    {\setlength{\fboxsep}{0pt}%
    \setlength{\fboxrule}{0.5pt}
    \fbox{\includegraphics[width=0.8\linewidth]{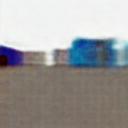}}} \\
    {\setlength{\fboxsep}{0pt}%
    \setlength{\fboxrule}{0.5pt}
    \fbox{\includegraphics[width=0.8\linewidth]{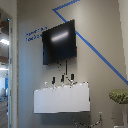}}}&
    {\setlength{\fboxsep}{0pt}%
    \setlength{\fboxrule}{0.5pt}
    \fbox{\includegraphics[width=0.8\linewidth]{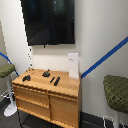}}}&
    {\setlength{\fboxsep}{0pt}%
    \setlength{\fboxrule}{0.5pt}
    \fbox{\includegraphics[width=0.8\linewidth]{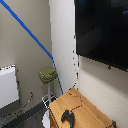}}}&
    {\setlength{\fboxsep}{0pt}%
    \setlength{\fboxrule}{0.5pt}
    \fbox{\includegraphics[width=0.8\linewidth]{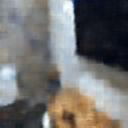}}}\\
    \end{tabularx}
    \caption{FDN image generation results. The first two columns are the inputs: first one provides the state information and the second provides the perspective information. The third column is the ground truth target image and the last column is the prediction from FDN.
    }
    \label{fig:fdn_results}
\end{figure}

\paragraph{Validation of Feature Disentanglement Network}
We present in Figure~\ref{fig:fdn_results} the results of image generation by composing state and perspective features using FDN. As illustrated in the generated image, the feature extraction network can compose state and perspective features to generate an image that has the same correspondence as the input state and perspective features.
In particular, the difference in the perspectives lies in the camera vertical position in simulation and camera vertical and horizontal location in real robot data. The results show that the network is able to learn such perspective information from training FDN. 

\begin{figure*}[t!]
\centering
\begin{tabularx}{\linewidth}{Y Y Y Y Y Y Y Y Y Y Y Y}
    \vspace*{0.05in}
    \tabincell{c}{$I_0^{robot}$} & \vspace*{0.05in} \tabincell{c}{$I_1^{robot}$} & \vspace*{0.05in} \tabincell{c}{$I_2^{robot}$} & \vspace*{0.05in} \tabincell{c}{$I_3^{robot}$} & 
    \vspace*{0.05in}\tabincell{c}{$I_4^{robot}$} & \vspace*{0.05in} \tabincell{c}{$I_5^{robot}$} &
    \vspace*{0.05in}\tabincell{c}{$I_6^{robot}$} & \vspace*{0.05in}\tabincell{c}{$I_7^{robot}$} &
    \vspace*{0.05in}\tabincell{c}{$I_8^{robot}$} &
    \vspace*{0.05in}\tabincell{c}{$I_9^{robot}$} &
    \vspace*{0.05in}\tabincell{c}{$I_{10}^{robot}$} & \vspace*{0.05in}\tabincell{c}{$I_g^{human}$}\\
    
    {\setlength{\fboxsep}{0pt}%
    \setlength{\fboxrule}{0.5pt}
    \fbox{\includegraphics[width=\linewidth,height=\linewidth]{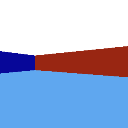}}} &
    {\setlength{\fboxsep}{0pt}%
    \setlength{\fboxrule}{0.5pt}
    \fbox{\includegraphics[width=\linewidth,height=\linewidth]{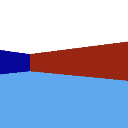}}} &
    {\setlength{\fboxsep}{0pt}%
    \setlength{\fboxrule}{0.5pt}
    \fbox{\includegraphics[width=\linewidth,height=\linewidth]{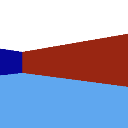}}} &
    {\setlength{\fboxsep}{0pt}%
    \setlength{\fboxrule}{0.5pt}
    \fbox{\includegraphics[width=\linewidth,height=\linewidth]{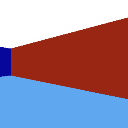}}} &
    {\setlength{\fboxsep}{0pt}%
    \setlength{\fboxrule}{0.5pt}
    \fbox{\includegraphics[width=\linewidth,height=\linewidth]{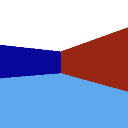}}} &
    {\setlength{\fboxsep}{0pt}%
    \setlength{\fboxrule}{0.5pt}
    \fbox{\includegraphics[width=\linewidth,height=\linewidth]{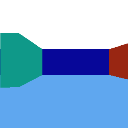}}} &
    {\setlength{\fboxsep}{0pt}%
    \setlength{\fboxrule}{0.5pt}
    \fbox{\includegraphics[width=\linewidth,height=\linewidth]{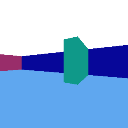}}} &
    {\setlength{\fboxsep}{0pt}%
    \setlength{\fboxrule}{0.5pt}
    \fbox{\includegraphics[width=\linewidth,height=\linewidth]{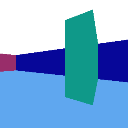}}} &
    {\setlength{\fboxsep}{0pt}%
    \setlength{\fboxrule}{0.5pt}
    \fbox{\includegraphics[width=\linewidth,height=\linewidth]{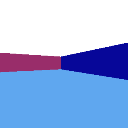}}} &
    {\setlength{\fboxsep}{0pt}%
    \setlength{\fboxrule}{0.5pt}
    \fbox{\includegraphics[width=\linewidth,height=\linewidth]{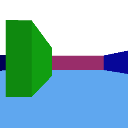}}} &
    {\setlength{\fboxsep}{0pt}%
    \setlength{\fboxrule}{0.5pt}
    \fbox{\includegraphics[width=\linewidth,height=\linewidth]{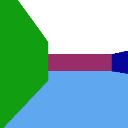}}} &
    {\setlength{\fboxsep}{0pt}%
    \setlength{\fboxrule}{0.5pt}
    \fbox{\includegraphics[width=\linewidth,height=\linewidth]{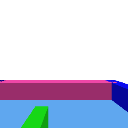}}} \\
    {\setlength{\fboxsep}{0pt}%
    \setlength{\fboxrule}{0.5pt}
    \fbox{\includegraphics[width=\linewidth,height=\linewidth]{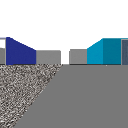}}} &
    {\setlength{\fboxsep}{0pt}%
    \setlength{\fboxrule}{0.5pt}
    \fbox{\includegraphics[width=\linewidth,height=\linewidth]{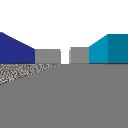}}} &
    {\setlength{\fboxsep}{0pt}%
    \setlength{\fboxrule}{0.5pt}
    \fbox{\includegraphics[width=\linewidth,height=\linewidth]{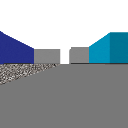}}} &
    {\setlength{\fboxsep}{0pt}%
    \setlength{\fboxrule}{0.5pt}
    \fbox{\includegraphics[width=\linewidth,height=\linewidth]{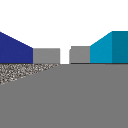}}} &
    {\setlength{\fboxsep}{0pt}%
    \setlength{\fboxrule}{0.5pt}
    \fbox{\includegraphics[width=\linewidth,height=\linewidth]{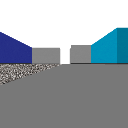}}} &
    {\setlength{\fboxsep}{0pt}%
    \setlength{\fboxrule}{0.5pt}
    \fbox{\includegraphics[width=\linewidth,height=\linewidth]{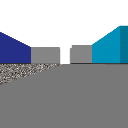}}} &
    {\setlength{\fboxsep}{0pt}%
    \setlength{\fboxrule}{0.5pt}
    \fbox{\includegraphics[width=\linewidth,height=\linewidth]{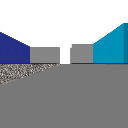}}} &
    {\setlength{\fboxsep}{0pt}%
    \setlength{\fboxrule}{0.5pt}
    \fbox{\includegraphics[width=\linewidth,height=\linewidth]{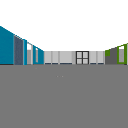}}} &
    {\setlength{\fboxsep}{0pt}%
    \setlength{\fboxrule}{0.5pt}
    \fbox{\includegraphics[width=\linewidth,height=\linewidth]{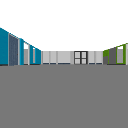}}} &
    {\setlength{\fboxsep}{0pt}%
    \setlength{\fboxrule}{0.5pt}
    \fbox{\includegraphics[width=\linewidth,height=\linewidth]{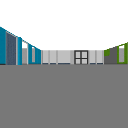}}} &
    {\setlength{\fboxsep}{0pt}%
    \setlength{\fboxrule}{0.5pt}
    \fbox{\includegraphics[width=\linewidth,height=\linewidth]{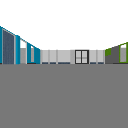}}} &
    {\setlength{\fboxsep}{0pt}%
    \setlength{\fboxrule}{0.5pt}
    \fbox{\includegraphics[width=\linewidth,height=\linewidth]{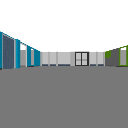}}} \\
    {\setlength{\fboxsep}{0pt}%
    \setlength{\fboxrule}{0.5pt}
    \fbox{\includegraphics[width=\linewidth,height=\linewidth]{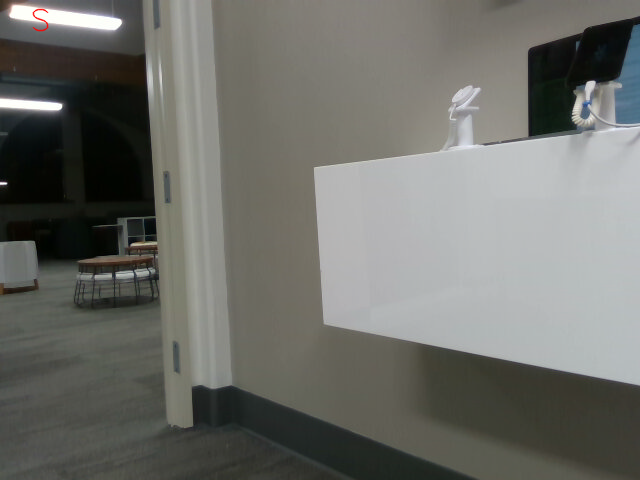}}} &
    {\setlength{\fboxsep}{0pt}%
    \setlength{\fboxrule}{0.5pt}
    \fbox{\includegraphics[width=\linewidth,height=\linewidth]{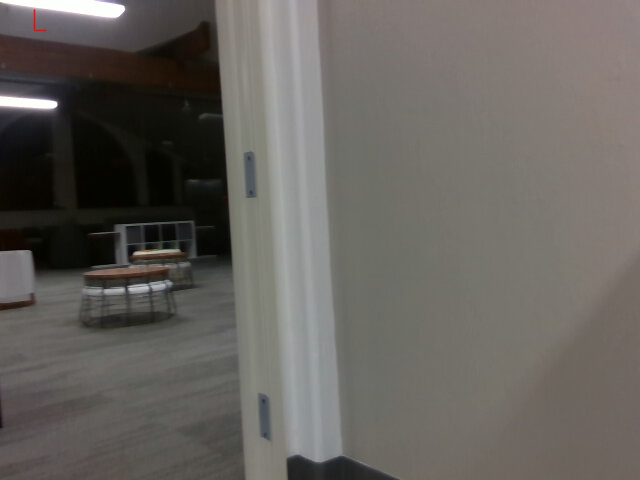}}} &
    {\setlength{\fboxsep}{0pt}%
    \setlength{\fboxrule}{0.5pt}
    \fbox{\includegraphics[width=\linewidth,height=\linewidth]{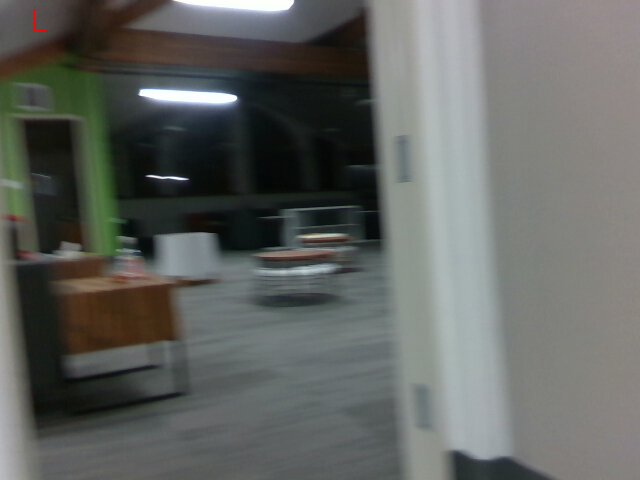}}} &
    {\setlength{\fboxsep}{0pt}%
    \setlength{\fboxrule}{0.5pt}
    \fbox{\includegraphics[width=\linewidth,height=\linewidth]{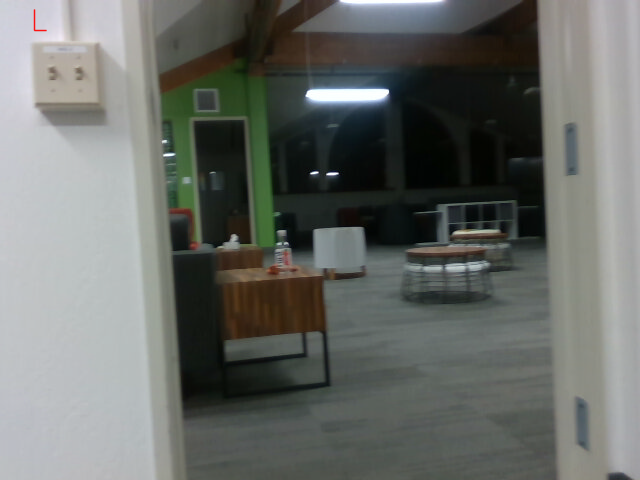}}} &
    {\setlength{\fboxsep}{0pt}%
    \setlength{\fboxrule}{0.5pt}
    \fbox{\includegraphics[width=\linewidth,height=\linewidth]{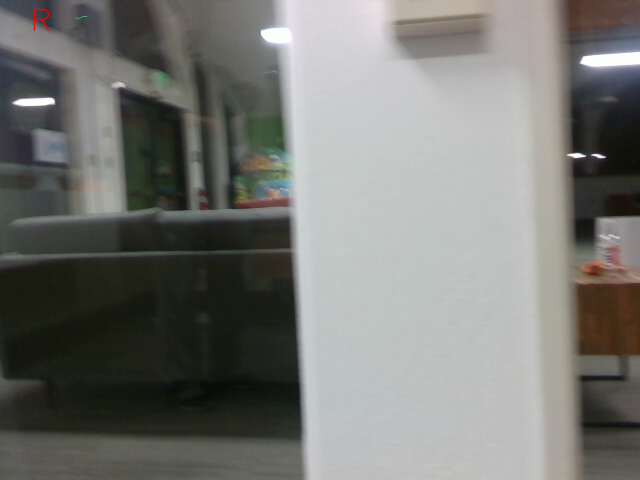}}} &
    {\setlength{\fboxsep}{0pt}%
    \setlength{\fboxrule}{0.5pt}
    \fbox{\includegraphics[width=\linewidth,height=\linewidth]{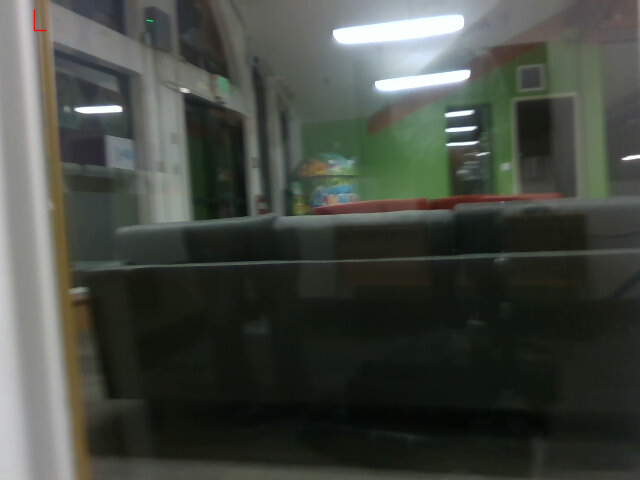}}} &
    {\setlength{\fboxsep}{0pt}%
    \setlength{\fboxrule}{0.5pt}
    \fbox{\includegraphics[width=\linewidth,height=\linewidth]{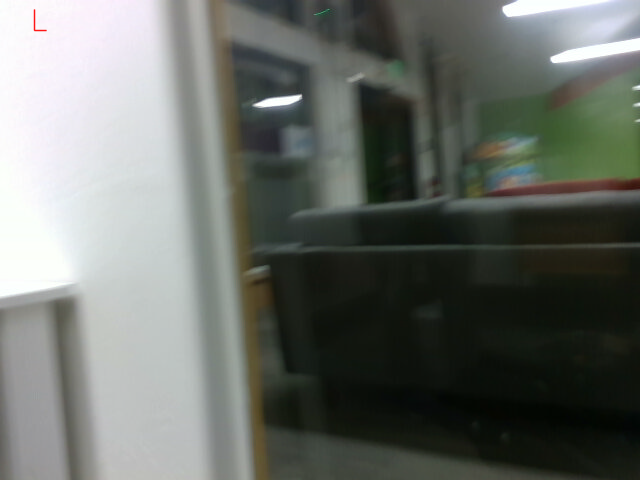}}} &
    {\setlength{\fboxsep}{0pt}%
    \setlength{\fboxrule}{0.5pt}
    \fbox{\includegraphics[width=\linewidth,height=\linewidth]{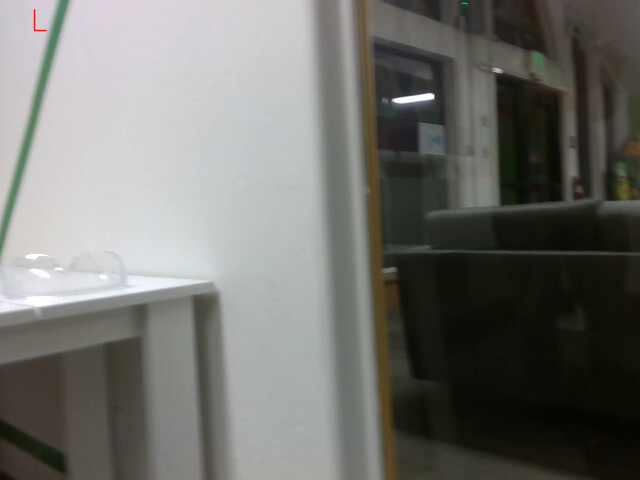}}} &
    {\setlength{\fboxsep}{0pt}%
    \setlength{\fboxrule}{0.5pt}
    \fbox{\includegraphics[width=\linewidth,height=\linewidth]{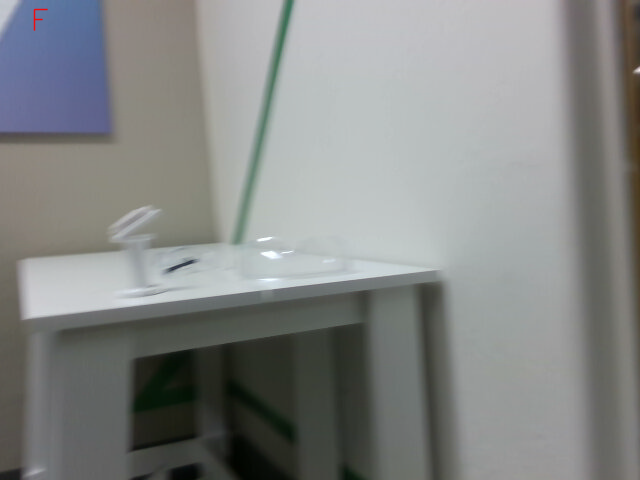}}} &
    {\setlength{\fboxsep}{0pt}%
    \setlength{\fboxrule}{0.5pt}
    \fbox{\includegraphics[width=\linewidth,height=\linewidth]{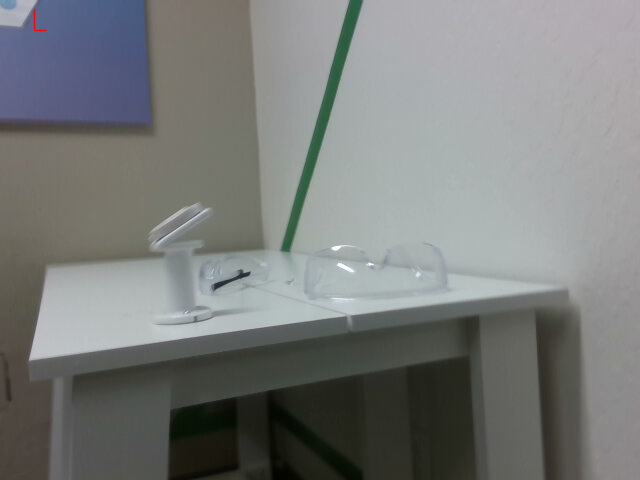}}} &
    {\setlength{\fboxsep}{0pt}%
    \setlength{\fboxrule}{0.5pt}
    \fbox{\includegraphics[width=\linewidth,height=\linewidth]{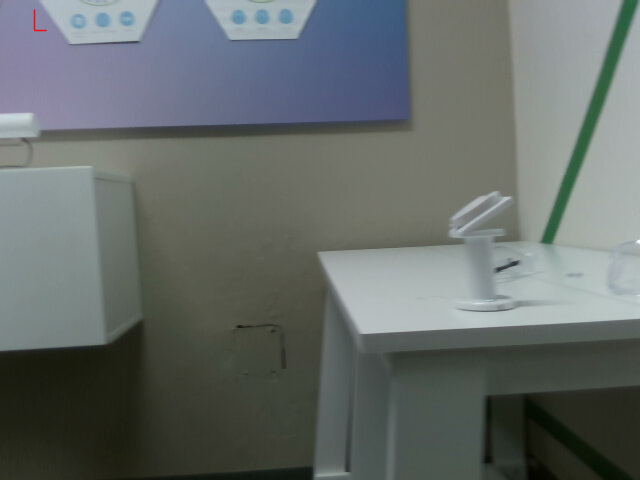}}} &
    {\setlength{\fboxsep}{0pt}%
    \setlength{\fboxrule}{0.5pt}
    \fbox{\includegraphics[width=\linewidth,height=\linewidth]{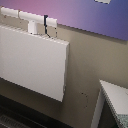}}} \\
    \end{tabularx}
    \caption{Robot successful visual navigation trajectories. In each row, the final image is the goal image and the first to the last second image show the robot navigation trajectory from the start to the goal location. Though captured from different perspectives, the goal specified by the human is the same as the last state of the robot.}
    \label{fig:success_traj_real}
    \vspace{-0.2in}
\end{figure*}

\begin{table}[t!]
    \vspace*{0.05in}
    \caption{Results with different task difficulties.}
    \centering
    \begin{tabular}{|c|c|c|c|}
        \hline
         Env/Distance & 2 & 5 & 10   \\
         \hline
         NavWorld & 79.66\% & 56.52\% & 21.81\% \\
         \hline
         OfficeWorld & 78.71\% & 54.00\% & 26.67\%\\
         \hline
    \end{tabular}
    \label{tab:pred_length}
\end{table}

\begin{table}[t!]
    \vspace*{0.05in}
    \caption{Results with different amount of human demonstrations. 
    }
    \centering
    \begin{tabular}{|c|c|c|c|}
        \hline
         Env/$\#$ of Human Demos & 100 & 300 & 500  \\
         \hline
         NavWorld & 21.81\% & 23.42\% & 25.23\% \\
         \hline
         OfficeWorld & 26.67\% & 31.00\% & 32.00\% \\
         \hline
    \end{tabular}
    \label{tab:success_rate1}
\end{table}
\paragraph{Simulation Results}
First, we validate our framework in the simulated environment. Our framework shows a learned zero-shot robot visual navigation behavior from human demonstrations with a success rate ranging from $20$\% to $80$\% depending on the task difficulty (see Table~\ref{tab:pred_length} for more details). We observe that a robot is able to find the goal position, specified from human's perspective, even when it is out of sight, indicating that the trained imitation learning model already models the environment with human demonstrations.

We evaluate the effect of task difficulty (by varying the number of minimum steps between the start and goal location) and the number of human demonstrations on the success rate.
For the latter, we fix the task difficulty to be 10 steps.
Table~\ref{tab:pred_length} shows that with more distance between the start and goal location, finding the correct path towards the goal becomes harder.
In a larger OfficeWorld environment, we observe such decrease in success rate with increasing task difficulty. Table~\ref{tab:success_rate1} shows that with more demonstrations the success rate indeed increases.

\paragraph{Hardware Results}
We provide results on the real legged robot, Laikago. 
We test the robot on three sets of goal-driven navigation tasks. We consider three targets and start locations to evaluate the robustness and consistency of the policy. The distance from the target location to the start location of the robot is around two meters. For each testing start and goal location pair, we test for three times and evaluate the success rate of the three trials. On these testing tasks, we obtain a success rate of around 60\%. We show one of the successful goal-driven visual navigation trajectories in Figure~\ref{fig:success_traj_real}. In our experience, a robot shows a better accuracy when the goal image is visually salient, such as a brown chair; it struggles to reach an object of colors similar to the background, e.g. a white desk in front of a white wall.
\vspace{-0.09in} 

\subsection{Analysis}
\begin{table}[t!]
    \vspace*{0.05in}
    \caption{Comparison of different feature loss on success rate.}
    \centering
    \begin{tabular}{|c|c|c|c|}
        \hline
         Success Rate/Loss & Ours & Cycle-triplet & Triplet  \\
         \hline
         NavWorld, 4 Steps &  54.55\% & 52.63\% & 8.77\% \\
         \hline
         NavWorld, 10 Steps & 21.81\% & 20.72\% & 2.70\% \\
         \hline
         OfficeWorld, 10 Steps & 26.67 \% & 27.30 \% & 25.08\% \\
         \hline
    \end{tabular}
    \label{tab:loss_compare}
\end{table}
\paragraph{Comparison with Different Loss Functions}
To investigate whether our proposed cycle-loss is suitable for training feature disentanglement, we compare with other baseline loss functions.
Specifically, we experiment with several combinations of the proposed cycle-loss and triplet loss~\cite{sermanet2018time}. In the first scenario, we train the feature extractor with our cycle-loss. In the second case, we combine cycle-loss with triplet loss. Given three images $I_{i}^p, I_i^q, I_j^p$, where $I_i^p$ and $I_i^q$ share their states and $I_i^p, I_j^p$ share their perspective, the triplet loss can be defined as, 
$L_{triplet}(\theta, I_i^p, I^q_i, I_j^p) =  \|F_{\theta}(I_i^p)-F_{\theta}(I_i^q)\|-\|F_{\theta}(I_i^p)-F_{\theta}(I_j^p)\|+\alpha.
    $
which minimizes state feature difference for the same state but from different perspectives, and maximize state feature difference for different states but from the same perspective. Here $\alpha$ is the enforced margin typically used in triplet-loss and usually the loss is cut to zero when it is negative. 
In the third case, we use triplet loss only to learn the state feature representation. 
The results are presented in Table~\ref{tab:loss_compare}.
It is clear that the Triplet loss alone has consistently worse results than our proposed Cycle loss. By combining Cycle-loss with Triplet loss the performance improved a bit compared to the Triplet loss. 
The triplet loss's poor performance may be a result of sensitivity to the data sampling process.
Our proposed Cycle-loss training is more stable and is not sensitive to data sampling.  Besides, triplet loss only learns the state feature (perspective-invariant feature) and our network learns both state and perspective feature and our decoder helps to verify the learned feature has correct correspondence in terms of states and perspectives.

\begin{table}[t!]
    \caption{Comparison of success rate with UPN and UPN-PerspChange.}
    \centering
    \begin{tabular}{|c|c|c|c|}
        \hline
         Env/Methods & Ours & UPN & UPN-PerspChange  \\
         \hline
         NavWorld & 54.55\% & 58.00\% & 7.88\% \\
         \hline
         OfficeWorld & 26.67\% & 27.00\% & 22.00\% \\
         \hline
    \end{tabular}
    \label{tab:upn}
\end{table}

\paragraph{Comparison with baselines}
We compare the proposed framework with a baseline algorithm, Universal Planning Networks (UPN)~\cite{srinivas2018universal}.
In particular, we test UPN in two scenarios: without and with perspective changes between a learner and a demonstrator. The former serves as the upper bound of our method's performance.
We will call the first method UPN and the second method UPN-PerspChange. In the first method UPN, we train and test UPN under the same perspective. In the second case, UPN is trained with multiple perspective data while the training perspective does not include the testing perspective. 

We perform a comparative study on NavWorld with 4 steps of start-goal distance and on OfficeWorld with 10 steps of start-goal distance. We choose these numbers of steps since one step in the environment can be a huge distance and the robot can move in a few steps to a state that is not visible from the starting state. The results are presented in Table~\ref{tab:upn}. 
The success rate of our method is approaching UPN, the empirical upper bound, indicating that FDN effectively handles the perspective shift.
When there is perspective change, UPN-PerspChange trained with some perspective data can't generalize to another unseen perspective, and the result is worse than our method. This indicates that the perspective shift is nontrivial and direct transfer does not work. 

We observe that the UPN-PerspChange works better in the OfficeWorld environment than in NavWorld, this is because that in the OfficeWorld environment, the turning angle is 90 degrees while in NavWorld the turning angle is 30 degrees. Therefore, to reach a state in the OfficeWorld, most of the actions are either moving forward or back. Even though the texture in the OfficeWorld is more complicated, the task difficulty with the same number of step between start and goal location is smaller. 

\paragraph{Scalability of the Proposed Approach} Though we rely on a GUI to label human data as an alternative of learning the robot inverse dynamics model, the real challenge in the Laikago robot is the inaccurate robot action labels, which results in inaccurate inverse dynamics model and thus making it less feasible to use the inverse dynamics model to label human data. To improve the scalability of the current framework and the success rate of goal driven visual navigation, we can improve the low-level control of the robot such that the behavior of the robot is more controllable. In addition, since we don't need to have very accurate human action labels (only in an abstract way indicating in which direction the human is going), optical flow~\cite{singh2006optical} methods can be used to automatically label human data.
\vspace{-0.08in} 

\section{Conclusion and Remarks}
We propose a novel imitation learning framework for learning a visual navigation policy on a legged robot from human demonstrations. Since it is hard to collect human data from the robot's perspective, one major challenge is to interpret the human demonstrations from different perspectives. To this end, we develop a feature disentanglement network (FDN) that extracts perspective-invariant features and a model-based imitation learning algorithm that trains a policy in the learned feature space. We demonstrate that the proposed framework can find effective navigation policies in two simulated worlds and one real environment. We further validate the framework by conducting ablation and comparative studies.

The bottleneck for deploying the current framework to real-world scenarios is the manual action labeling process of human demonstrations. However, automated action labeling is not straightforward at the required high accuracy ($> 90$\%). One possible approach is to collect a small amount of the robot’s navigation data to build an inverse dynamics model that takes in two consecutive images and predicts the robot’s action. In our experience, this approach works in simulation but not on the real robot because a legged robot’s gait blurs the camera images. In addition, the robot’s discrete actions are often not well-matched with real human demonstrations. In the future, we want to investigate more stable gaits with continuous control commands.

Although we tested the framework for learning a legged robot policy from human demonstrations, the framework is designed to support general imitation learning between any heterogeneous agents. In the future, we hope to build a general system that can learn navigation policies for data-expensive robots, such as legged robots or aerial vehicles, from easy-to-operate robots, such as mobile robots, autonomous cars or humans. If we can fully exploit a large navigation data sets, such as Google Streetview~\cite{anguelov2010google}, there is great potential to significantly improve the performance on real robots.

\bibliography{reference}
\bibliographystyle{IEEEtranS}

\end{document}








%% file: root.bbl
\begin{thebibliography}{10}
\providecommand{\url}[1]{#1}
\csname url@samestyle\endcsname
\providecommand{\newblock}{\relax}
\providecommand{\bibinfo}[2]{#2}
\providecommand{\BIBentrySTDinterwordspacing}{\spaceskip=0pt\relax}
\providecommand{\BIBentryALTinterwordstretchfactor}{4}
\providecommand{\BIBentryALTinterwordspacing}{\spaceskip=\fontdimen2\font plus
\BIBentryALTinterwordstretchfactor\fontdimen3\font minus
  \fontdimen4\font\relax}
\providecommand{\BIBforeignlanguage}[2]{{%
\expandafter\ifx\csname l@#1\endcsname\relax
\typeout{** WARNING: IEEEtranS.bst: No hyphenation pattern has been}%
\typeout{** loaded for the language `#1'. Using the pattern for}%
\typeout{** the default language instead.}%
\else
\language=\csname l@#1\endcsname
\fi
#2}}
\providecommand{\BIBdecl}{\relax}
\BIBdecl

\bibitem{anguelov2010google}
D.~Anguelov, C.~Dulong, D.~Filip, C.~Frueh, S.~Lafon, R.~Lyon, A.~Ogale,
  L.~Vincent, and J.~Weaver, ``Google street view: Capturing the world at
  street level,'' \emph{Computer}, vol.~43, no.~6, pp. 32--38, 2010.

\bibitem{bonin2008visual}
F.~Bonin-Font, A.~Ortiz, and G.~Oliver, ``Visual navigation for mobile robots:
  A survey,'' \emph{Journal of intelligent and robotic systems}, vol.~53,
  no.~3, pp. 263--296, 2008.

\bibitem{chiang2019learning}
H.-T.~L. Chiang, A.~Faust, M.~Fiser, and A.~Francis, ``Learning navigation
  behaviors end-to-end with autorl,'' \emph{IEEE Robotics and Automation
  Letters}, vol.~4, no.~2, pp. 2007--2014, 2019.

\bibitem{codevilla2018end}
F.~Codevilla, M.~Miiller, A.~L{\'o}pez, V.~Koltun, and A.~Dosovitskiy,
  ``End-to-end driving via conditional imitation learning,'' in \emph{2018 IEEE
  International Conference on Robotics and Automation (ICRA)}.\hskip 1em plus
  0.5em minus 0.4em\relax IEEE, 2018, pp. 1--9.

\bibitem{coumans2016pybullet}
E.~Coumans and Y.~Bai, ``Pybullet, a python module for physics simulation for
  games, robotics and machine learning,'' \emph{GitHub repository}, 2016.

\bibitem{dijkstra1959note}
E.~W. Dijkstra, ``A note on two problems in connexion with graphs,''
  \emph{Numerische mathematik}, vol.~1, no.~1, pp. 269--271, 1959.

\bibitem{fang2019scene}
K.~Fang, A.~Toshev, L.~Fei-Fei, and S.~Savarese, ``Scene memory transformer for
  embodied agents in long-horizon tasks,'' in \emph{Proceedings of the IEEE
  Conference on Computer Vision and Pattern Recognition}, 2019, pp. 538--547.

\bibitem{francis2019prmrl}
A.~Francis, A.~Faust, H.~L. Chiang, J.~Hsu, J.~C. Kew, M.~Fiser, and T.~E. Lee,
  ``Long-range indoor navigation with {PRM-RL},'' \emph{CoRR}, vol.
  abs/1902.09458, 2019.

\bibitem{gupta2017cognitive}
S.~Gupta, J.~Davidson, S.~Levine, R.~Sukthankar, and J.~Malik, ``Cognitive
  mapping and planning for visual navigation,'' in \emph{Proceedings of the
  IEEE Conference on Computer Vision and Pattern Recognition}, 2017, pp.
  2616--2625.

\bibitem{huber1992robust}
P.~J. Huber, ``Robust estimation of a location parameter,'' in
  \emph{Breakthroughs in statistics}.\hskip 1em plus 0.5em minus 0.4em\relax
  Springer, 1992, pp. 492--518.

\bibitem{jha2018disentangling}
A.~H. Jha, S.~Anand, M.~Singh, and V.~Veeravasarapu, ``Disentangling factors of
  variation with cycle-consistent variational auto-encoders,'' in
  \emph{European Conference on Computer Vision}.\hskip 1em plus 0.5em minus
  0.4em\relax Springer, 2018, pp. 829--845.

\bibitem{fu2019language}
F.~Justin, K.~Anoop, L.~Sergey, and G.~Sergio, ``From language to goals:
  Inverse reinforcement learning for vision-based instruction following,'' in
  \emph{7th International Conference on Learning Representations, {ICLR}},
  2019.

\bibitem{kerl2013dense}
C.~Kerl, J.~Sturm, and D.~Cremers, ``Dense visual slam for rgb-d cameras,'' in
  \emph{2013 IEEE/RSJ International Conference on Intelligent Robots and
  Systems}.\hskip 1em plus 0.5em minus 0.4em\relax IEEE, 2013, pp. 2100--2106.

\bibitem{kingma2014adam}
D.~P. Kingma and J.~Ba, ``Adam: {A} method for stochastic optimization,'' in
  \emph{3rd International Conference on Learning Representations, {ICLR} 2015},
  Y.~Bengio and Y.~LeCun, Eds., 2015.

\bibitem{konolige2010view}
K.~Konolige, J.~Bowman, J.~Chen, P.~Mihelich, M.~Calonder, V.~Lepetit, and
  P.~Fua, ``View-based maps,'' \emph{The International Journal of Robotics
  Research}, vol.~29, no.~8, pp. 941--957, 2010.

\bibitem{kumar2019learning}
A.~Kumar, S.~Gupta, and J.~Malik, ``Learning navigation subroutines by watching
  videos,'' in \emph{Conference on Robot Learning}, 2019.

\bibitem{lee2018diverse}
H.-Y. Lee, H.-Y. Tseng, J.-B. Huang, M.~Singh, and M.-H. Yang, ``Diverse
  image-to-image translation via disentangled representations,'' in
  \emph{Proceedings of the European conference on computer vision (ECCV)},
  2018, pp. 35--51.

\bibitem{lenz2015deepmpc}
I.~Lenz, R.~A. Knepper, and A.~Saxena, ``Deepmpc: Learning deep latent features
  for model predictive control.'' in \emph{Robotics: Science and
  Systems}.\hskip 1em plus 0.5em minus 0.4em\relax Rome, Italy, 2015.

\bibitem{liu2018imitation}
Y.~Liu, A.~Gupta, P.~Abbeel, and S.~Levine, ``Imitation from observation:
  Learning to imitate behaviors from raw video via context translation,'' in
  \emph{2018 IEEE International Conference on Robotics and Automation
  (ICRA)}.\hskip 1em plus 0.5em minus 0.4em\relax IEEE, 2018, pp. 1118--1125.

\bibitem{meltzoff1999born}
A.~N. Meltzoff, ``Born to learn: What infants learn from watching us,''
  \emph{The role of early experience in infant development}, pp. 145--164,
  1999.

\bibitem{pan2019semantic}
X.~Pan, X.~Chen, Q.~Cai, J.~Canny, and F.~Yu, ``Semantic predictive control for
  explainable and efficient policy learning,'' in \emph{2019 International
  Conference on Robotics and Automation (ICRA)}.\hskip 1em plus 0.5em minus
  0.4em\relax IEEE, 2019, pp. 3203--3209.

\bibitem{pan2018agile}
Y.~Pan, C.-A. Cheng, K.~Saigol, K.~Lee, X.~Yan, E.~Theodorou, and B.~Boots,
  ``Agile autonomous driving using end-to-end deep imitation learning,'' in
  \emph{Robotics: science and systems}, 2018.

\bibitem{pathak2018zero}
D.~Pathak, P.~Mahmoudieh, G.~Luo, P.~Agrawal, D.~Chen, Y.~Shentu, E.~Shelhamer,
  J.~Malik, A.~A. Efros, and T.~Darrell, ``Zero-shot visual imitation,'' in
  \emph{Proceedings of the IEEE Conference on Computer Vision and Pattern
  Recognition Workshops}, 2018, pp. 2050--2053.

\bibitem{raibert1986legged}
M.~H. Raibert, \emph{Legged robots that balance}.\hskip 1em plus 0.5em minus
  0.4em\relax MIT press, 1986.

\bibitem{ramachandran2017swish}
P.~Ramachandran, B.~Zoph, and Q.~V. Le, ``Swish: a self-gated activation
  function,'' \emph{arXiv preprint arXiv:1710.05941}, vol.~7, 2017.

\bibitem{rhinehart2018deep}
N.~Rhinehart, R.~McAllister, and S.~Levine, ``Deep imitative models for
  flexible inference, planning, and control,'' \emph{arXiv preprint
  arXiv:1810.06544}, 2018.

\bibitem{ross2011reduction}
S.~Ross, G.~Gordon, and D.~Bagnell, ``A reduction of imitation learning and
  structured prediction to no-regret online learning,'' in \emph{Proceedings of
  the fourteenth international conference on artificial intelligence and
  statistics}, 2011, pp. 627--635.

\bibitem{sermanet2018time}
P.~Sermanet, C.~Lynch, Y.~Chebotar, J.~Hsu, E.~Jang, S.~Schaal, S.~Levine, and
  G.~Brain, ``Time-contrastive networks: Self-supervised learning from video,''
  in \emph{2018 IEEE International Conference on Robotics and Automation
  (ICRA)}.\hskip 1em plus 0.5em minus 0.4em\relax IEEE, 2018, pp. 1134--1141.

\bibitem{singh2006optical}
S.~P. Singh, P.~J. Csonka, and K.~J. Waldron, ``Optical flow aided motion
  estimation for legged locomotion,'' in \emph{2006 IEEE/RSJ International
  Conference on Intelligent Robots and Systems}.\hskip 1em plus 0.5em minus
  0.4em\relax IEEE, 2006, pp. 1738--1743.

\bibitem{srinivas2018universal}
A.~Srinivas, A.~Jabri, P.~Abbeel, S.~Levine, and C.~Finn, ``Universal planning
  networks: Learning generalizable representations for visuomotor control,'' in
  \emph{Proceedings of the 35th International Conference on Machine Learning,
  {ICML}}, vol.~80, 2018, pp. 4739--4748.

\bibitem{stadie2017third}
B.~C. Stadie, P.~Abbeel, and I.~Sutskever, ``Third person imitation learning,''
  in \emph{5th International Conference on Learning Representations, {ICLR}},
  2017.

\bibitem{laikago}
\BIBentryALTinterwordspacing
Unitree. (2019) Laikago. [Online]. Available:
  \url{http://www.unitree.cc/e/action/ShowInfo.php?classid=6&id=1}
\BIBentrySTDinterwordspacing

\bibitem{wang2019reinforced}
X.~Wang, Q.~Huang, A.~Celikyilmaz, J.~Gao, D.~Shen, Y.-F. Wang, W.~Y. Wang, and
  L.~Zhang, ``Reinforced cross-modal matching and self-supervised imitation
  learning for vision-language navigation,'' in \emph{Proceedings of the IEEE
  Conference on Computer Vision and Pattern Recognition}, 2019, pp. 6629--6638.

\bibitem{yu2018bdd100k}
F.~Yu, W.~Xian, Y.~Chen, F.~Liu, M.~Liao, V.~Madhavan, and T.~Darrell,
  ``Bdd100k: A diverse driving video database with scalable annotation
  tooling,'' \emph{arXiv preprint arXiv:1805.04687}, 2018.

\bibitem{zhu2017unpaired}
J.-Y. Zhu, T.~Park, P.~Isola, and A.~A. Efros, ``Unpaired image-to-image
  translation using cycle-consistent adversarial networks,'' in
  \emph{Proceedings of the IEEE international conference on computer vision},
  2017, pp. 2223--2232.

\bibitem{zhu2017visual}
Y.~Zhu, D.~Gordon, E.~Kolve, D.~Fox, L.~Fei-Fei, A.~Gupta, R.~Mottaghi, and
  A.~Farhadi, ``Visual semantic planning using deep successor
  representations,'' in \emph{Proceedings of the IEEE International Conference
  on Computer Vision}, 2017, pp. 483--492.

\bibitem{zhu2017target}
Y.~Zhu, R.~Mottaghi, E.~Kolve, J.~J. Lim, A.~Gupta, L.~Fei-Fei, and A.~Farhadi,
  ``Target-driven visual navigation in indoor scenes using deep reinforcement
  learning,'' in \emph{2017 IEEE international conference on robotics and
  automation (ICRA)}.\hskip 1em plus 0.5em minus 0.4em\relax IEEE, 2017, pp.
  3357--3364.

\bibitem{ziebart2008maximum}
B.~D. Ziebart, A.~L. Maas, J.~A. Bagnell, and A.~K. Dey, ``Maximum entropy
  inverse reinforcement learning.'' in \emph{AAAI}, vol.~8.\hskip 1em plus
  0.5em minus 0.4em\relax Chicago, IL, USA, 2008, pp. 1433--1438.

\end{thebibliography}
